\documentclass[letterpaper]{article} 
\usepackage{aaai2026}  
\usepackage{times}  
\usepackage{helvet}  
\usepackage{courier}  
\usepackage[hyphens]{url}  
\usepackage{graphicx} 
\usepackage{natbib}  
\usepackage{caption} 
\usepackage{algorithm}
\usepackage{algorithmic}

\usepackage{newfloat}
\usepackage{listings}
\DeclareCaptionStyle{ruled}{labelfont=normalfont,labelsep=colon,strut=off} 
\floatstyle{ruled}
\newfloat{listing}{tb}{lst}{}
\floatname{listing}{Listing}
\usepackage{lipsum}
\usepackage{graphicx}
\usepackage{algorithm}
\usepackage{algorithmic}
\usepackage{nicematrix,booktabs}
\usepackage{multirow}
\usepackage{ragged2e}
\usepackage{tabularx}
\usepackage{tcolorbox}
\usepackage{subfigure}
\usepackage{transparent}
\usepackage{listings}
\usepackage{boldline}
\usepackage{arydshln}
\usepackage{diagbox}
\usepackage{amssymb}
\usepackage{pifont}

\usepackage{tikz}
\usepackage{enumitem}
\usepackage{changepage}
\usepackage{multirow}
\newcommand{\m}{UNBench}

\title{Benchmarking LLMs for Political Science: A United Nations Perspective}

\author{
    Yueqing Liang\textsuperscript{\rm 1}, 
    Liangwei Yang\textsuperscript{\rm 2}, 
    Chen Wang\textsuperscript{\rm 3}, 
    Congying Xia\textsuperscript{\rm 4}, 
    Rui Meng\textsuperscript{\rm 2}\thanks{Now at Google Cloud AI Research.}, \\
    Xiongxiao Xu\textsuperscript{\rm 1}, 
    Haoran Wang\textsuperscript{\rm 6}, 
    Ali Payani\textsuperscript{\rm 5}, 
    Kai Shu\textsuperscript{\rm 6}\thanks{Corresponding author}
}
\affiliations{
    \textsuperscript{\rm 1}Illinois Institute of Technology\quad
    \textsuperscript{\rm 2}Salesforce Research \quad
    \textsuperscript{\rm 3}University of Illinois Chicago\\
    \textsuperscript{\rm 4}Meta GenAI\quad
    \textsuperscript{\rm 5}Cisco Research\quad
    \textsuperscript{\rm 6}Emory University\\
    \{yliang40, xxu85\}@hawk.illinoistech.edu, \{liangwei.yang, ruimeng\}@salesforce.com, cwang266@uic.edu, congyingxia@meta.com,    apayani@cisco.com, \{haoran.wang, kai.shu\}@emory.edu
}

\begin{document}

\thispagestyle{plain}

\makeatletter
\def\ps@plain{%
  \def\@oddfoot{\hfil\thepage\hfil}%
  \def\@evenfoot{\hfil\thepage\hfil}%
}
\pagestyle{plain}
\makeatother

\maketitle

\begin{abstract}
Large Language Models (LLMs) have achieved significant advances in natural language processing, yet their potential for high-stakes political decision-making remains largely unexplored. This paper addresses the gap by focusing on the application of LLMs to the United Nations (UN) decision-making process, where the stakes are particularly high and political decisions can have far-reaching consequences. We introduce a novel dataset comprising publicly available UN Security Council (UNSC) records from 1994 to 2024, including draft resolutions, voting records, and diplomatic speeches. Using this dataset, we propose the United Nations Benchmark (UNBench), the first comprehensive benchmark designed to evaluate LLMs across four interconnected political science tasks: co-penholder judgment, representative voting simulation, draft adoption prediction, and representative statement generation. These tasks span the three stages of the UN decision-making process—drafting, voting, and discussing—and aim to assess LLMs' ability to understand and simulate political dynamics. Our experimental analysis demonstrates the potential and challenges of applying LLMs in this domain, providing insights into their strengths and limitations in political science. To the best of our knowledge, this is the first benchmark to systematically evaluate LLMs in UN decision-making, contributing to the growing intersection of AI and political science.

\end{abstract}

\begin{links}
    \link{UNBench}{https://github.com/yueqingliang1/UNBench}
    \link{Extended version}{https://arxiv.org/abs/2502.14122}
\end{links}

\section{Introduction}

Large Language Models (LLMs) such as GPT-4~\cite{openai2023gpt4}, Llama~\cite{dubey2024llama}, and DeepSeek~\cite{liu2024deepseek} have achieved unprecedented proficiency in language tasks and are increasingly under development tailored for different domains~\cite{cheng2023adapting}. Yet, their adaptation to high-stakes political decision-making remains underexplored—particularly in scenarios where model outputs could influence real-world governance. Political science demands capabilities beyond semantic understanding: predicting coalition dynamics, interpreting ambiguous diplomatic language, and navigating the tension between national interests and global norms. These challenges unfold within the United Nations (UN), where a single draft resolution, once adopted, becomes binding international law under Chapter VII of the UN Charter, with cascading impacts on global security, trade, and human rights (e.g., Resolution 1973’s no-fly zone over Libya in 2011).
Studying the application of LLMs in political science represents both a technical frontier and a critical societal challenge.

\begin{figure}[t]
    \centering
    \includegraphics[width=1.0\linewidth]{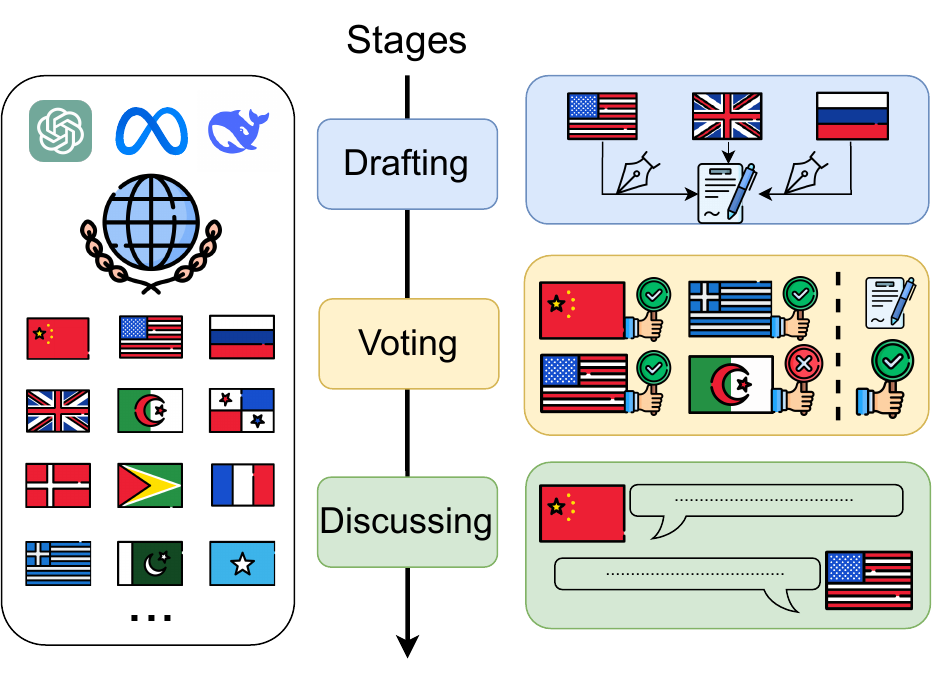}
    \caption{Three key stages of the United Nations decision-making, where LLMs can simulate an individual member.}
    \label{fig:scenario}
\end{figure}

The scope of UN resolutions is extensive, extending far beyond political statements. Adopted resolutions can authorize military interventions (such as Resolution 678 in 1990), impose sanctions that cripple national economies (e.g., Resolution 1718 on North Korea in 2006), or redefine global priorities (e.g., Resolution 2341 on critical infrastructure protection in 2017). 
By analyzing the open-access data from draft resolutions and meeting records, we can explore how current LLMs understand the critical issues facing the international community and assess their ability to interpret bilateral and multilateral relations. This extends LLM's application toward political science, enhancing the analysis of international policies and diplomacy.

Currently, there is no comprehensive benchmark designed specifically for LLM applications in political science. Existing benchmarks (e.g., MMLU~\cite{hendryckstest2021,hendrycks2021ethics}, BIG-Bench~\cite{suzgun2022challenging}) with related tasks remain fragmented, and their designs may not adequately reflect LLMs' understanding of political science.
Such fragmented evaluation overlooks the interconnected nature of real-world political decision-making, particularly in high-stakes multilateral settings like the UN. 
Figure~\ref{fig:scenario} shows the stages of each UN draft resolution. It consists of three stages. 
(1) Drafting: The creation of the resolution's text, involving the collaboration among member states.
(2) Voting: The process in which the resolution is formally adopted or rejected by voting from the UN members.
(3) Discussing: Each member states the rationality of their voting. 
Different tasks occur in different stages and also interconnected with each other.

To fill the benchmark gap in political science, we introduce \textbf{United Nations Benchmark (UNBench)}, the first comprehensive benchmark to evaluate LLMs' ability across four distinct yet interconnected political science tasks of different stages:
(a) \textbf{Co-Penholder Judgement}: Given anonymized draft content, identify optimal co-author nations, simulating coalition-building strategies in multilateral diplomacy.
(b) \textbf{Representatives Voting Simulation}: Instruct an LLM to act as a national agent (e.g., "As the U.S. representative") and output voting decisions (['In favour', 'Against', etc.]), testing contextual understanding of national interests.
(c) \textbf{Draft Adoption Prediction}: Input a draft resolution to LLM and ask it to predict the draft adoption probability, which requires analysis of historical voting patterns and geopolitical alignments.
(d) \textbf{Representative Statement Generation}: Generate country-specific speeches justifying voting positions, evaluating persuasive language generation under political constraints.
The tasks are designed from different UN stages, varying across predictive and generative tasks.
Our benchmark is built from publicly available UNSC official records (1994-2024), comprising draft resolutions, voting records, and diplomatic speeches which are extracted from meeting records. In summary, our work makes the following contributions:

\begin{itemize}
    \item Systematically curated and processed United Nations data from 1994 to 2024, including draft resolutions, voting records, and diplomatic speeches, to provide a new and comprehensive dataset that facilitates LLM applications in political science.
    \item Introduce the first comprehensive benchmark in political science, designed to assess LLM performance across various real-world tasks in different stages of the UN decision-making process.
    \item Conduct extensive experimental analysis on the benchmark, demonstrating the effectiveness and limitations of current LLMs in handling complex political tasks.
\end{itemize}

\section{Related Works}

\subsection{LLMs in Political Science}

Existing benchmark datasets have played a critical role in advancing Large Language Models (LLMs) in political science tasks. Datasets such as OpinionQA~\cite{santurkar2023whose}, PerSenT~\cite{bastan2020author}, and GermEval-2017~\cite{wojatzki2017germeval} evaluate LLMs on classifying sentiments or identifying topics within political texts. These benchmarks primarily emphasize static text understanding. BillSum~\cite{kornilova2019billsum} and CaseLaw~\cite{shu2024lawllm} specialize in summarization or analysis of legislative documents. Datasets like PolitiFact~\cite{shu2020fakenewsnet}, GossipCop~\cite{grover2022public}, and Weibo~\cite{jin2017multimodal} focus on detecting misinformation. Election prediction and voting behavior datasets, such as U.S. Senate Returns 2020~\cite{DVN/ER9XTV_2022} and State Precinct-Level Returns 2018~\cite{DVN/ZFXEJU_2022} evaluate models' capabilities in statistical pattern recognition and forecasting. These tasks often rely on structured data and focus on predictive performance related to electoral outcomes. Our benchmark is the first to unify these diverse political science tasks within a single, end-to-end benchmark grounded in real-world multilateral decision-making process in the United Nations scenario.

\subsection{United Nations-Related Datasets}

The United Nations has long been a focal institution for global politics, attracting extensive study in political science ~\cite{bailey2017estimating, voeten2013data}. While various publicly available datasets shed partial light on UN processes, they typically focus on limited aspects of the organization. Harvard Dataverse UN Voting Dataset~\cite{DVN/LEJUQZ_2009} compiles pairwise country voting statistics, offering quantitative insights but lacking textual data such as draft resolutions or debate transcripts. UNSCR.com~\cite{unscr2025} collects mainly \textit{adopted} Security Council resolutions, providing topic labels and related resolutions but minimal coverage of \textit{draft} content. UNSCdeb8~\cite{unscdeb8_zenodo} includes verbatim debate transcripts from 2010 to 2017, capturing real-time deliberation but omitting links to draft texts or voting outcomes. Lastly, the UN Parallel Corpus~\cite{un_parallel_corpus} supplies multilingual final resolutions and meeting records (1994–2014), yet lacks draft-stage materials and detailed metadata to trace evolving negotiations. In this paper, we present the first LLM benchmark covering the full UN resolution process—combining draft texts, debates, and voting records from 1994–2024 to capture the complete decision-making trajectory.

\section{The United Nations Benchmark}
\label{Sec:Method}

\begin{figure*}[t]
\centering
\includegraphics[width=1.0\linewidth]{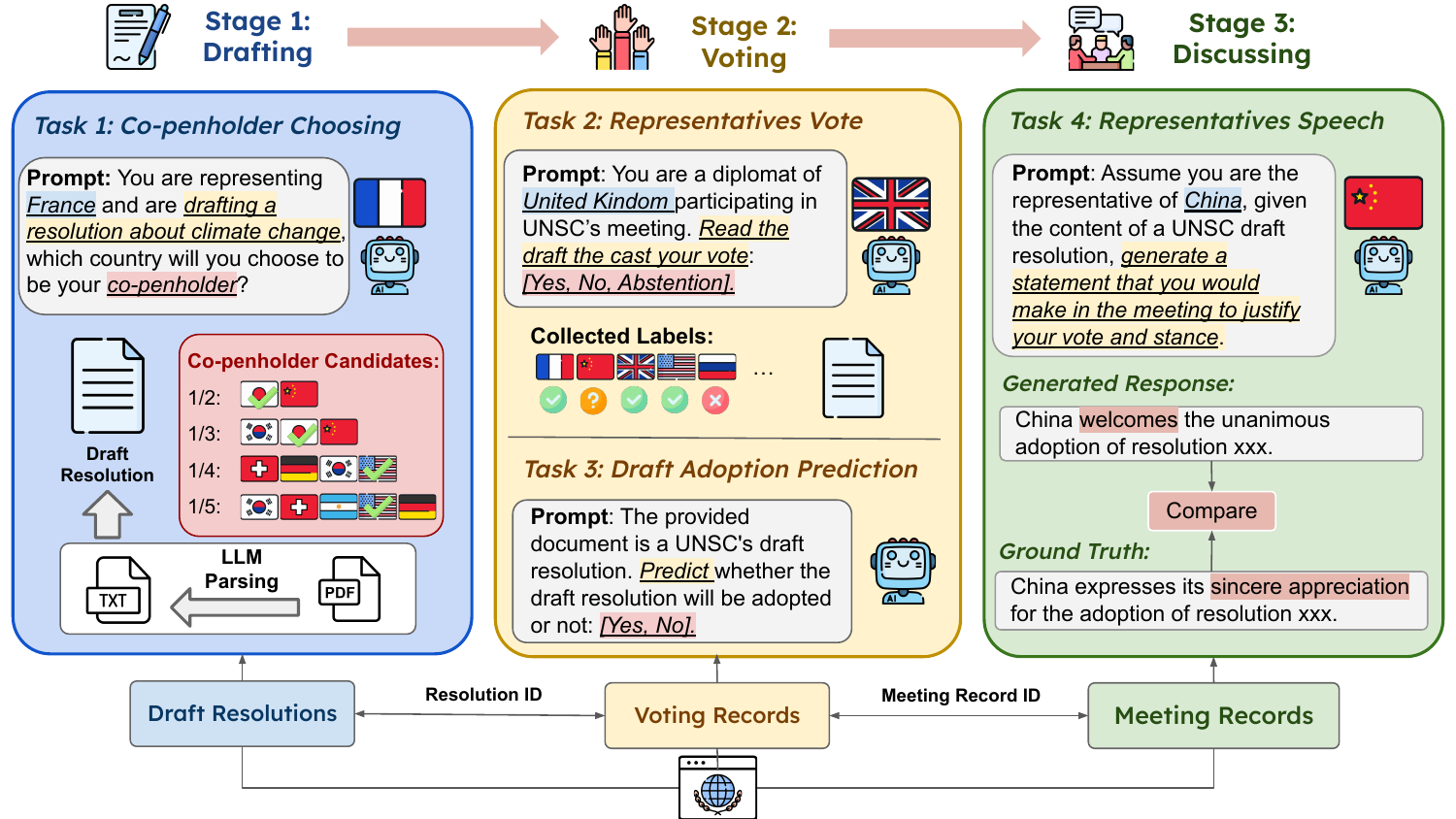}
\caption{The proposed \textbf{{\m}}. It consists of $4$ tasks extracted from different stages of a UN draft.}
\label{fig:major}
\end{figure*}

UNBench is extracted from United Nations resolution decisions. It consists of four tasks, covering three distinct stages (Drafting, Voting, Discussing) before the adoption/rejection of each resolution.

\subsection{Benchmark Notation Definition}
To formalize our benchmark and the tasks, we introduce the following notations:

\begin{itemize}
    \item \(\mathcal{R} = \{r_{1}, r_{2}, \dots, r_{N}\}\): The set of all \textbf{draft resolutions}. Each resolution \(r_{i}\) contains proposed actions, mandates, and contextual details (e.g., sponsoring countries).
    
    \item \(\mathcal{C} = \{c_{1}, c_{2}, \dots, c_{M}\}\): The set of all \textbf{UN members}, including both permanent and non-permanent members.
    
    \item For each resolution \(r_{i} \in \mathcal{R}\), \(\mathcal{C}_{\mathrm{candidate}}(r_{i}) \subseteq \mathcal{C}\) denotes the \textbf{candidate co-penholder} countries—those likely to sponsor or support \(r_{i}\) during the \emph{drafting} process.
    
    \item \(\mathcal{V}(r_{i}) = \{v_{i,1}, v_{i,2}, \dots, v_{i,15}\}\): The \textbf{votes} of the 15 Security Council members on \(r_{i}\). Each vote \(v_{i,j}\) is one of \(\{\texttt{[In Favour]}, \texttt{[Against]}, \texttt{[Abstention]}\}\).
    
    \item \(\text{Result}(r_{i}) \in \{\textsc{Adopted}, \textsc{NotAdopted}\}\): The \textbf{decision} on \(r_{i}\). It is \(\textsc{NotAdopted}\) if it fails to secure the necessary majority or is vetoed by a permanent member.
    
    \item \(\mathcal{S}(r_{i}, c_{j})\): The \textbf{official statement} (speech) of country \(c_{j}\) regarding \(r_{i}\), delivered during the \emph{discussion} stage. This typically includes the country’s rationale, policy concerns, and diplomatic stance.
\end{itemize}

\subsection{Stage 1: Drafting}
\label{sec:drafting}

Drafting is the initial phase in the lifecycle of a resolution. Typically, one or more countries—often referred to as \emph{penholders}—take the lead in preparing a draft text, outlining the resolution’s objectives, scope, and operative clauses. The draft is then refined through closed-door consultations, circulated informally among Council members. A unique feature of drafting is the practice of \emph{co-penholdership}, wherein multiple countries jointly sponsor or “own” the resolution from its inception. We design a task focusing on \emph{identifying the most suitable co-penholder}.

\paragraph{Task 1: Co-Penholder Judgement.}  
Formally, let \(r_{i}\) be a draft resolution authored by country \(c_{a}\). We sample \(\mathcal{C}_{\mathrm{candidate}}(r_{i}) \subseteq \mathcal{C}\) as a set of co-penholders candidates, each representing a different country. The LLM is prompted to assume the \emph{role of the author country} \(c_{a}\), given the text of \(r_{i}\), and asked to choose exactly one co-penholder from the set \(\mathcal{C}_{\mathrm{candidate}}(r_{i})\). In practice, we vary the number of candidates from 2 to 5, making this task a multi-choice setup with a single correct answer.

Co-penholdership reflects shared strategic interests, diplomatic partnerships, or specialized expertise on the issue at hand. Identifying a suitable co-penholder requires the model to have the following ability:

\begin{itemize}
    \item \textbf{Comprehend Contextual Information:} Understand the resolution’s key themes (e.g., conflict prevention, sanctions regime, peacekeeping mandates), and recognize which policy domains (e.g., human rights, climate, nuclear disarmament) are relevant. This tests the model’s ability to integrate textual comprehension of policy content with broader geopolitical and diplomatic reasoning.

    \item \textbf{Infer Diplomatic Alignments:} Analyze historical or implied alignments and identify country pairings likely to co-sponsor a resolution. This evaluates whether the model can correlate textual cues with knowledge of past collaborations or alliances, and navigate multi-choice questions where the differences between candidate countries may be subtle or context-dependent.

    \item \textbf{Reason About Multilateral Cooperation:} Weigh factors such as a candidate country’s veto power (if permanent), geopolitical priorities, and regional interests to recommend a co-penholder that maximizes the resolution’s likelihood of success.
    
\end{itemize}

Hence, Task~1 offers a focused measure of the model’s capacity to perform political and textual reasoning in a controlled, multi-choice format, laying the groundwork for subsequent stages involving voting and post-vote deliberation.

\subsection{Stage 2: Voting}
\label{sec:voting}

In the second stage of the resolution lifecycle, each of the 15 Council members casts a vote to determine whether a draft resolution is \emph{adopted} or \emph{rejected}. Permanent members wield veto power, meaning a single ``\texttt{Against}'' vote from any of the five permanent countries can block the resolution, regardless of overall support. Non-permanent members, on the other hand, primarily influence the outcome through collective consensus and persuasive negotiation. Based on this voting mechanism, we define two tasks that capture different facets of decision-making at this stage.

\paragraph{Task 2: Representatives Voting Simulation.}
Formally, for a draft resolution \(r_{i}\), let \(\mathcal{V}(r_{i}) = \{v_{i,1}, \ldots, v_{i,15}\}\) denote the votes cast by each of the 15 Council members. In this task, the LLM is given the content of \(r_{i}\) and prompted to \emph{assume the role of a specific representative} \(c_{j}\) (where \(c_{j} \in \mathcal{C}\)) to determine how that country would vote on \(r_{i}\). Each vote \(v_{i,j}\) must be one of \(\{\texttt{[In Favour]}, \texttt{[Against]}, \texttt{[Abstention]}\}\).

\paragraph{Objective and Challenges.}
Effective voting simulation requires the model to:
\begin{itemize}
    \item \textbf{Comprehend the Resolution:} Interpret the text of \(r_{i}\) in light of its subject matter (e.g., conflict prevention).
    \item \textbf{Incorporate National Interests:} Weigh a representative country’s known priorities and geopolitical alignments (e.g., historical alliances, regional blocs).
    \item \textbf{Account for Veto Power:} Recognize whether \(c_{j}\) is a permanent member with veto ability.
\end{itemize}
These dimensions test not only the model’s text understanding but also its ability in political and strategic reasoning, reflecting real-world complexities in UN negotiations.

\paragraph{Task 3: Draft Adoption Prediction.}
Once the votes \(\mathcal{V}(r_{i})\) are cast, the resolution is \emph{adopted} if it secures the necessary majority (i.e., at least nine \texttt{[In Favour]} votes) and no permanent member exercises a veto. In this task, the LLM receives the text of \(r_{i}\) and \emph{predict the final outcome}, denoted
\[
    \mathrm{Result}(r_{i}) \;\in\;\{\textsc{Adopted}, \textsc{NotAdopted}\}.
\]
Unlike Task~2, which focuses on individual country votes, this task tests whether the model can account for the collective dynamics of all 15 Council members. Key factors include but are not limited to overall council sentiment, potential veto threats, historical precedents, etc.
Accurate adoption prediction thus demands higher-level inference about the distribution of possible votes, the interplay of veto power, and the delicate balancing of geopolitical interests. Together, Tasks~2 and~3 provide complementary perspectives on an LLM’s capacity to model real-world decision-making under complex international relations.

\subsection{Stage 3: Discussing}
\label{sec:discussing}

Once the voting concludes, each member typically delivers a statement clarifying the vote and articulating national positions or broader policy perspectives. These statements reveal the rationale behind each country’s stance—whether \emph{In Favour}, \emph{Against}, or \emph{Abstention}. Since these statements are given in an open discussion format, countries may engage in debates, directly addressing or countering the arguments made by other members. This final discussion phase can shape diplomatic narratives surrounding the resolution’s implications and signal future policy directions. We design a task that evaluates an LLM’s ability to generate representative statements aligned with national interests, voting behavior, and diplomatic discourse norms.

\paragraph{Task 4: Representative Statement Generation}
Formally, for a draft resolution \(r_{i}\), let \(\mathcal{S}(r_{i}, c_{j})\) denote the official statement made by country \(c_{j}\). In this task, the LLM receives the text of \(r_{i}\) alongside contextual details, including the outcome of the vote, each country's voting decision, and any prior statements made in the discussion (if available, in the order they were delivered). The model is then asked to \emph{generate the statement} that \(c_{j}\) would deliver. This statement should reflect:

\begin{itemize}
    \item \textbf{National Interests and Policies:} How does \(c_{j}\)’s geopolitical position shape its response to the resolution (e.g., security concerns, regional dynamics)?
    \item \textbf{Vote Justification:} If \(c_{j}\) voted \texttt{[In Favour]}, \texttt{[Against]}, or \texttt{[Abstention]}, the statement should provide a coherent rationale for the decision.
    \item \textbf{Diplomatic Tone and Style:} UNSC discourse follows a formal, measured tone. The model should generate text that aligns with the conventions of diplomatic statements.
\end{itemize}

By prompting the model to produce \emph{country-specific} statements, Task~4 evaluates higher-level language generation skills in a multi-faceted political context. The ability to incorporate historical alliances, policy priorities, and rhetorical conventions into coherent and persuasive statements indicates an advanced understanding of both textual composition and global political dynamics.

\section{Dataset Construction}
\label{sec:data_construct}

Our dataset \(\mathcal{D}\) is constructed from United Nations Security Council (UNSC) meeting records, draft resolutions, and voting histories spanning the years 1994 to 2024. The resulting corpus not only includes the textual content of each draft resolution but also contextual metadata such as voting outcomes, sponsoring nations, meeting transcripts, and the temporal sequence of events.

The overarching goal of this work is to provide a \emph{unified and extensive} collection of the decision-making process, thereby enabling evaluation of multiple LLM capabilities in a single benchmark. To achieve this, we collect multi-perspective data from the official website and digital library\cite{UNSC_digital_lib}, which archive draft resolutions, voting records, and meeting minutes. Below, we highlight key challenges and our corresponding strategies in constructing \(\mathcal{D}\) in the different stages of our benchmark construction.

\paragraph{Data Collection.}
In the data collection stage, we have three challenges: 
(1) \textit{\textbf{Fragmented Records.}} Draft resolutions, voting logs, and meeting transcripts reside in separate sections of the UN database. We utilize shared identifiers (e.g., resolution numbers, meeting record IDs) to \emph{link} these sources. As illustrated schematically in Figure~\ref{fig:major}, we first retrieve all draft resolutions, then query corresponding voting records by resolution ID (when applicable), and finally map meeting transcripts via the meeting record ID.
(2) \textit{\textbf{Missing or Incomplete Metadata.}} Despite the UN’s comprehensive record-keeping, certain entries contain missing fields (e.g., sponsor lists), inconsistent data formats, or broken links. We mitigate these issues by cross-referencing multiple UN repositories, manually curating ambiguous entries, and applying standardized naming conventions for country references.
(3) \textit{\textbf{Historical Document Diversity.}} The official document formats and website structures vary considerably across decades, complicating automated crawling and parsing. We address this by implementing adaptive web-scraping scripts that detect layout differences and by performing iterative quality checks to ensure data consistency.

\paragraph{Data Conversion.}
UN documents are primarily stored in PDF format, making direct ingestion by current LLMs infeasible. We therefore extract and convert the content into plain text. Early attempts using generic Python PDF libraries yielded mixed accuracy due to the unstructured, domain-specific nature of political documents. We applied a \emph{LLM-based} parser (LlamaParse\cite{llamaparse}) to handle complex formatting (e.g., multi-column layouts, footnotes, multilingual text).

\paragraph{Data Processing.}
(1) \textit{\textbf{Labeling Adopted vs.\ Unadopted Drafts.}} Some drafts never become official resolutions (i.e., “unadopted”), lacking a formal resolution ID. We thus inspect the official notes in each draft’s record and cross-verify with the final resolution index to categorize them correctly.
(2) \textit{\textbf{Country Name Normalization.}} Different records refer to the same country with variations (e.g., “United Kingdom” vs.\ “Kingdom”). We employ the official name to unify references to the same country entity.
(3) \textit{\textbf{Metadata Alignment.}} For each draft \(r_i\), we compile the relevant information: author/sponsor countries, date, issue category, voting breakdown, and meeting transcripts into a structured format compatible with modern NLP frameworks.

Through these steps, UNBench incorporates the \emph{entire} lifecycle of each UNSC draft resolution, from initial sponsorship and negotiations to final votes and discussions, ensuring comprehensive coverage for our benchmark tasks.

\section{Experiments}
\label{Sec:Exp}

\subsection{Dataset Statistics}

Our {\m} covers a broad range of draft resolutions, voting records, and meeting transcripts, providing diverse scenarios for evaluating multiple LLM capabilities. As shown in Table~\ref{tab:dataset_stats}, Task~1 features 1{,}300 draft resolutions with a total of 355{,}126 instances—reflecting a multi-choice setup where each instance corresponds to an author country selecting a co-penholder. Task~2 contains 17{,}430 instances of individual votes for each country that participating in the voting, while Task~3 comprises 1{,}978 draft resolutions with both \texttt{[Adopted]} and \texttt{[NotAdopted]} labels. Finally, Task~4 includes 7{,}394 statements from 1{,}752 UNSC meetings, testing the ability to generate coherent speeches that align with national stances. More data analysis can be found at Appendix~A in the extended version \cite{liang2025benchmarking}.

\begin{table}[!htp]
\centering
\begin{tabularx}{\linewidth}{llr}
\toprule
\textbf{Task} & \textbf{Statistic} & \textbf{Value} \\
\midrule
\multirow{4}{*}{Task 1} 
    & \# Drafts                 & 1,300  \\
    & \# Unique Draft Authors   & 209    \\
    & Avg. \# Authors per Draft  & 7      \\
    & \# Total Instances         & 355,126\\
\midrule
\multirow{5}{*}{Task 2} 
    & \# Drafts                 & 1,162  \\
    & \# Total Instances         & 17,430 \\
    & \# \texttt{[In Favour]}    & 17,020 \\
    & \# \texttt{[Against]}      & 16     \\
    & \# \texttt{[Abstention]}   & 391    \\
\midrule
\multirow{3}{*}{Task 3} 
    & \# Drafts                 & 1,978  \\
    & \# \texttt{[Adopted]}      & 1,880  \\
    & \# \texttt{[NotAdopted]}   & 98     \\
\midrule
\multirow{4}{*}{Task 4} 
    & \# Meetings (Drafts)         & 1,752  \\
    & \# Statements              & 7,394  \\
    & \# Countries               & 204    \\
    & Avg. \# Tokens per Statement & 450  \\
\bottomrule
\end{tabularx}
\caption{Statistics for our {\m}.}
\label{tab:dataset_stats}
\end{table}

\subsection{Experimental Setup}

\paragraph{Models.}
Tasks 1, 2, and 3 are classification-oriented. We compare two \emph{traditional text classification models} (BERT~\cite{devlin2018bert} and DeBERTa~\cite{he2020deberta}) against several \emph{instruction-tuned LLMs}: \textit{Llama-3.2-1B-Instruct}~\cite{dubey2024llama}, \textit{Llama-3.2-3B-Instruct}, \textit{Llama-3.1-8B-Instruct}, \textit{Mistral-7B-Instruct}~\cite{jiang2023mistral}, \textit{DeepSeek-V3}~\cite{liu2024deepseek}, \textit{Qwen2.5-7B-Instruct}~\cite{yang2024qwen2}, and \textit{GPT-4o} via the Azure API. BERT and DeBERTa are fine-tuned for three epochs with a learning rate of \(5\times10^{-5}\). Llama models run on an 8\(\times\)A6000 GPU server, while Mistral and DeepSeek are accessed through the TogetherAI platform. Since Task~4 (representatives statement generation) is inherently a generative task, we only evaluate LLMs on it, setting a temperature of 0.0 for consistent comparisons and adjusting maximum output lengths per task.

\paragraph{Settings.}
For each classification-oriented task, we employ a time-based train/test split. Specifically, we reserve half of the samples from the less frequent labels (according to chronological order) as the test set, ensuring the training set remains balanced and temporally earlier. This protocol simulates real-world scenarios where future events must be predicted from past data.

\paragraph{Metrics.}
Task~1 is a multiple-choice question with \(k\) ranging from 2 to 5. We calculate accuracy by checking whether the model identifies the single valid co-penholder. Tasks~2 and~3 are classification problems (multi-class and binary, respectively), so we report metrics robust to class imbalance, including F1-score, balanced accuracy (Bal. ACC), and PR AUC. Task~4 is evaluated using text-generation metrics (ROUGE) and semantic similarity (Sentence-BERT~\cite{reimers2019sentence}) to measure how closely the generated statements match ground truth in style and content. For brevity, we present only two primary metrics per task in Table~\ref{tab:exp_major}, with detailed breakdowns available at Appendix~\ref{sec:appendix_results}. 

\begin{table*}[t]\centering
\resizebox{0.9\textwidth}{!}{%
\begin{tabular}{lp{1cm}p{1cm}p{1cm}p{1cm}p{1cm}p{1cm}p{1.1cm}p{1.1cm}}
\toprule
&\multicolumn{2}{c}{\textbf{Task 1}} &\multicolumn{2}{c}{\textbf{Task 2}} &\multicolumn{2}{c}{\textbf{Task 3}} &\multicolumn{2}{c}{\textbf{Task 4}} \\\cmidrule{2-9}
\textbf{Model} & (1/2) & (1/5) &Bal. ACC &PR AUC &Bal. ACC &Mac. F1 &ROUGE &Cosine Sim. \\\midrule
BERT &0.011 &0.010 &0.537 &0.396 &0.333 &0.328 &/ &/ \\

DeBERTa &0.010 &0.011 &0.500 &0.527 &0.333 &0.328 &/ &/ \\

Llama-3.2-1B &0.581 &0.269 &0.546 &0.185 &0.320 &0.326 &0.033 &0.329 \\

Llama-3.2-3B &0.578 &0.297 &0.597 &0.385 &0.597 &\textbf{0.402} &0.041 &0.290 \\

Llama-3.1-8B &0.665 &0.379 &0.530 &0.168 &0.357 &0.359 &0.039 &0.355 \\

Mistral-7B &0.563 &0.281 &0.426 &0.268 &0.529 &0.140 &0.194 &0.575 \\

GPT-4o &\textbf{0.726} &\textbf{0.464} &\textbf{0.823} &\textbf{0.696} &\textbf{0.677} &\underline{0.363} & 0.199 &\underline{0.619} \\

Qwen2.5-7B & 0.642 & 0.293 & 0.699 & 0.375 & 0.578 & 0.241 & \underline{0.201} & \textbf{0.623} \\

DeepSeek-V3 &\underline{0.695} &\underline{0.422} &\underline{0.724} &\underline{0.655} &\underline{0.668} &0.351 &\textbf{0.207} &\textbf{0.623} \\
\bottomrule
\end{tabular}
}
\caption{Our {\m} contains four tasks. For each task, we choose two metrics to show. (1/k) means choosing 1 from k choices, Bal. ACC is balance accuracy, PR AUC is precision-recall AUC. The best results for each metric are highlighted in \textbf{bold}, while the second-best results are \underline{underlined}. More results could be found at Appendix B.}
\label{tab:exp_major}
\end{table*}

\subsection{Task 1: Co-Penholder Judgement}

\begin{figure}[t]
    \centering
    \includegraphics[width=1.0\linewidth]{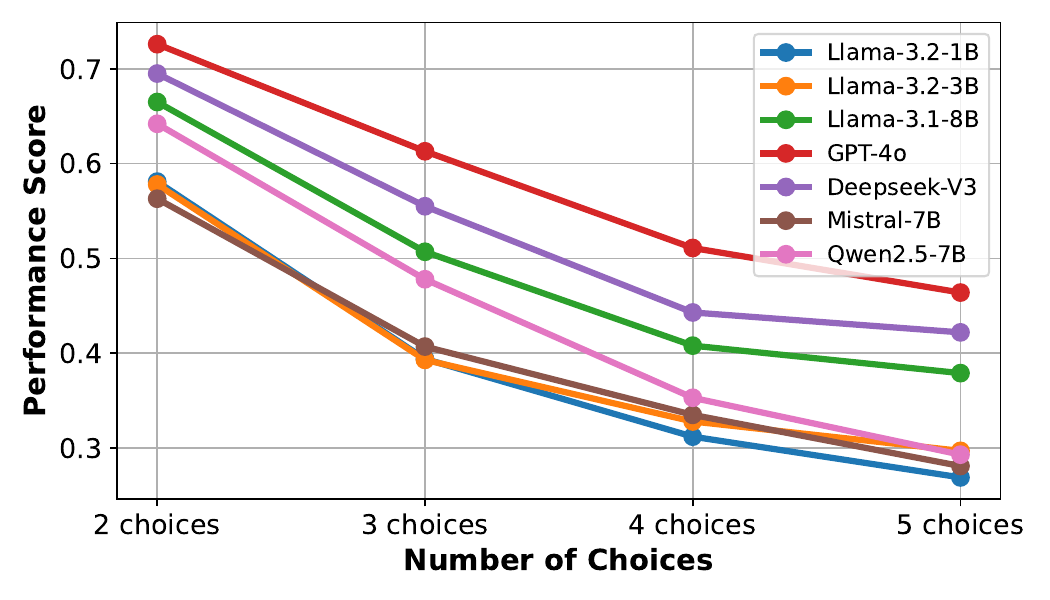}
    \caption{Models performance in Task 1 by varying the number of choices. }
    \label{fig:task1}
\end{figure}

This task evaluates LLMs' ability to \textit{\textbf{identify strategic geopolitical alliances}} by selecting co-penholders for UNSC draft resolutions, requiring nuanced understanding of international relations and procedural norms. GPT-4o (0.726) and DeepSeek-V3 (0.695) dominate, demonstrating superior contextual reasoning and geopolitical knowledge. Smaller LLMs (e.g., Llama-3.2-1B: 0.581) lag significantly, while traditional models (BERT: 0.011) fail entirely, underscoring the necessity of LLM-scale architectures for complex political inference. In addition, we vary the number of candidate choices (2–5) to test models’ robustness under increasing complexity. As shown in Figure~\ref{fig:task1}, all models exhibit declining accuracy as choices increase, with GPT-4o maintaining dominance across all levels. The widening performance gaps as choices increase highlight the divergent capacities of LLMs to resolve ambiguity, validating that large, modern LLMs excel at synthesizing latent political knowledge, while smaller or traditional models lack the representational capacity for such nuanced reasoning.

\subsection{Task 2: Representatives Voting Simulation}
Focused on simulating country-specific voting behavior, this task tests models’ ability to \textit{\textbf{infer issue-specific voting patterns}} by analyzing how nations prioritize resolution content and contextual geopolitical dynamics. Unlike Task 1, which evaluates proactive alliance-building, Task 2 emphasizes reactive decision-making based on the interplay of national interests, ideological alignment, and external pressures. GPT-4o achieves the highest performance (0.823 Bal. ACC, 0.696 PR AUC), demonstrating a strong ability to model nuanced trade-offs. DeepSeek-V3 (0.724 Bal. ACC, 0.655 PR AUC) and Llama-3.2-3B (0.597 Bal. ACC) show moderate success but struggle with ambiguous cases, while Mistral-7B (0.426 Bal. ACC) performs poorly, reflecting its inability to systematically weigh competing factors. The results highlight LLMs’ potential to simulate diplomatic behavior but reveal significant variance in their capacity to reason about issue-specific voting dynamics. 
We also assess whether model performance varies with respect to \textit{\textbf{recency}}, \textit{\textbf{potential geopolitical disparities}}, and different \textit{\textbf{contextual metadata}}. Detailed results are provided at Appendix C.1, C.2, and C.3.

\subsection{Task 3: Draft Adoption Prediction}
Whereas Task 2 centers on individual votes, Task 3 measures \textit{\textbf{document-level outcome prediction}}, requiring holistic reasoning about all 15 Council members to predict whether a resolution is eventually \texttt{[Adopted]} or \texttt{[NotAdopted]}. GPT-4o shows the best Bal. ACC (\(0.677\)) and a competitive macro-F1 (\(0.363\)), while Llama-3.2-3B surpasses others in macro-F1 (\(0.402\)) but has lower Bal. ACC (\(0.597\)). The divergence between Bal. ACC and F1 metrics reveals the challenge of modeling adoption mechanics, where understanding Council-wide political dynamics, potential veto threats, and support coalitions play roles. 
In addition, we analyze GPT-4o’s robustness across time, regional authorship, and metadata conditions for this task. The corresponding evaluations are detailed at Appendix C.1, C.2, and C.3.

\subsection{Task 4: Representatives Statement Generation}
This task evaluates LLMs’ ability to generate \textit{\textbf{style-sensitive diplomatic statements}} that align with country-specific rhetoric and protocol. Qwen2.5-7B and DeepSeek-V3 tie for semantic fidelity (0.623 Cosine), demonstrating strong alignment with the intended meaning and tone of diplomatic statements. DeepSeek-V3 also leads in lexical overlap (0.207 ROUGE), suggesting better adherence to precise terminological requirements. Mistral-7B achieves high Cosine similarity (0.575) but modest ROUGE (0.194), indicating strength in paraphrasing and conceptual alignment rather than verbatim replication. All models underperform in ROUGE, exposing limitations in precise terminological alignment—a critical requirement for diplomatic drafting. This highlights the unresolved challenge of balancing creativity and protocol adherence in LLM-generated diplomatic text, particularly in capturing the formal and nuanced language of international diplomacy.

\paragraph{Cross-Task Summary.}
Each task targets a distinct facet of UNSC decision-making. Task~1 primarily tests \textit{\textbf{textual and geopolitical reasoning}} in a multi-choice format, Task~2 and Task~3 emphasize \textit{\textbf{political prediction capabilities}} (from simulating individual votes to forecasting final outcomes), and Task~4 stresses \textit{\textbf{diplomatic language generation}}, requiring alignment with formal protocols and country-specific rhetoric. Performance gaps across tasks and models highlight both the promise and complexity of applying LLMs to real-world international governance, reinforcing the need for dedicated benchmarks as {\m}.

\section{Potential Applications}
\label{sec:potential_app }

The {\m} offers significant value to both LLM researchers and stakeholders in international governance, enabling practical applications and advancing research in geopolitical AI. Below, we outline potential use cases for different stakeholders.

\paragraph{For LLM Researchers:}
{\m} provides a rich testbed for advancing research in LLMs, particularly in the context of geopolitical reasoning and time-series analysis: (1) \textbf{\textit{Geopolitical Reasoning}}: The tasks in the benchmark span a wide range of capabilities, from alliance identification (Task~1) to issue-specific voting prediction (Task~2), offering researchers a comprehensive framework for evaluating and improving LLMs’ understanding of international relations. (2) \textbf{\textit{Temporal Analysis}}: With data spanning 30 years, {\m} enables time-series tasks such as predicting trends in diplomatic behavior, forecasting shifts in international alliances, or analyzing the impact of historical events (e.g., the end of the Cold War) on UNSC dynamics. For instance, researchers could use the dataset to predict how emerging global issues (e.g., climate change) will influence future resolutions. (3) \textbf{\textit{Fine-Grained Prediction}}: The benchmark’s focus on multi-choice and generative tasks challenges researchers to develop models that balance precision and creativity. For example, improving ROUGE scores in Task~4 could lead to breakthroughs in generating protocol-compliant diplomatic text. (4) \textbf{\textit{Bias Analysis}}: UNBench provides an opportunity to study biases in LLMs’ geopolitical reasoning, ensuring that models do not perpetuate stereotypes or oversimplify complex international dynamics.

\paragraph{For UN Stakeholders:}
The ability to predict and analyze UNSC decision-making using LLMs also has implications for diplomats, policymakers, and international organizations: (1) \textbf{\textit{Draft Adoption Forecasting}}: By predicting whether a draft resolution will be adopted (Task~3), stakeholders can proactively adjust negotiation strategies, allocate resources more effectively, and build coalitions to maximize the likelihood of success. 
For example, knowing that a climate resolution is unlikely to pass could prompt earlier lobbying efforts or revisions to the draft. 
(2) \textbf{\textit{Voting Behavior Simulation}}: Simulating country-specific voting behavior (Task~2) allows stakeholders to anticipate the positions of key nations, identify potential allies or opponents, and tailor diplomatic outreach accordingly. This could be particularly useful for smaller nations or NGOs seeking to navigate complex geopolitical landscapes.

\section{Conclusion and Future Work}
This paper introduces UNBench, the first comprehensive benchmark for evaluating LLMs' capabilities in political science through UN Security Council records (1994-2024). By designing four interconnected tasks spanning the complete UN resolution lifecycle, we provide a more authentic framework for assessing LLMs' understanding of complex diplomatic dynamics. It not only addresses the current gap in LLM evaluation frameworks but also establishes a foundation for future research between artificial intelligence and international relations, demonstrating how LLMs could potentially assist in analyzing global governance processes. More detailed analyses can be found at Appendix in the extended version \cite{liang2025benchmarking}.

Future work can extend {\m} to multilingual settings for evaluating models across diverse diplomatic languages. Researchers may also use {\m} to study more realistic multi-agent simulations where LLMs act as countries, as well as multimodal inputs that incorporate maps, timelines, or historical context. Another direction is Constructing richer country profiles to enhance grounding and strategic consistency when models simulate state behavior.

\section*{Acknowledgments}

This material is based upon work supported by NSF awards (SaTC-2241068, IIS-2506643, and POSE-2346158), a Cisco Research Award, and a Microsoft Accelerate Foundation Models Research Award. The views and conclusions contained in this document are those of the authors and should not be interpreted as necessarily representing the official policies, either expressed or implied, of the National Science Foundation.

\bibliography{aaai2026}

\appendix

\label{sec:appendix}

\section{Data Analysis}
\label{sec:appendix_data_analysis}
This section introduces our data analysis on the collected datasets from United Nation.

\subsection{Subjects Analysis}

\subsubsection{Author-subject Co-occurrence Analysis}

\begin{figure*}[t]
\centering
\includegraphics[width=0.9\linewidth]{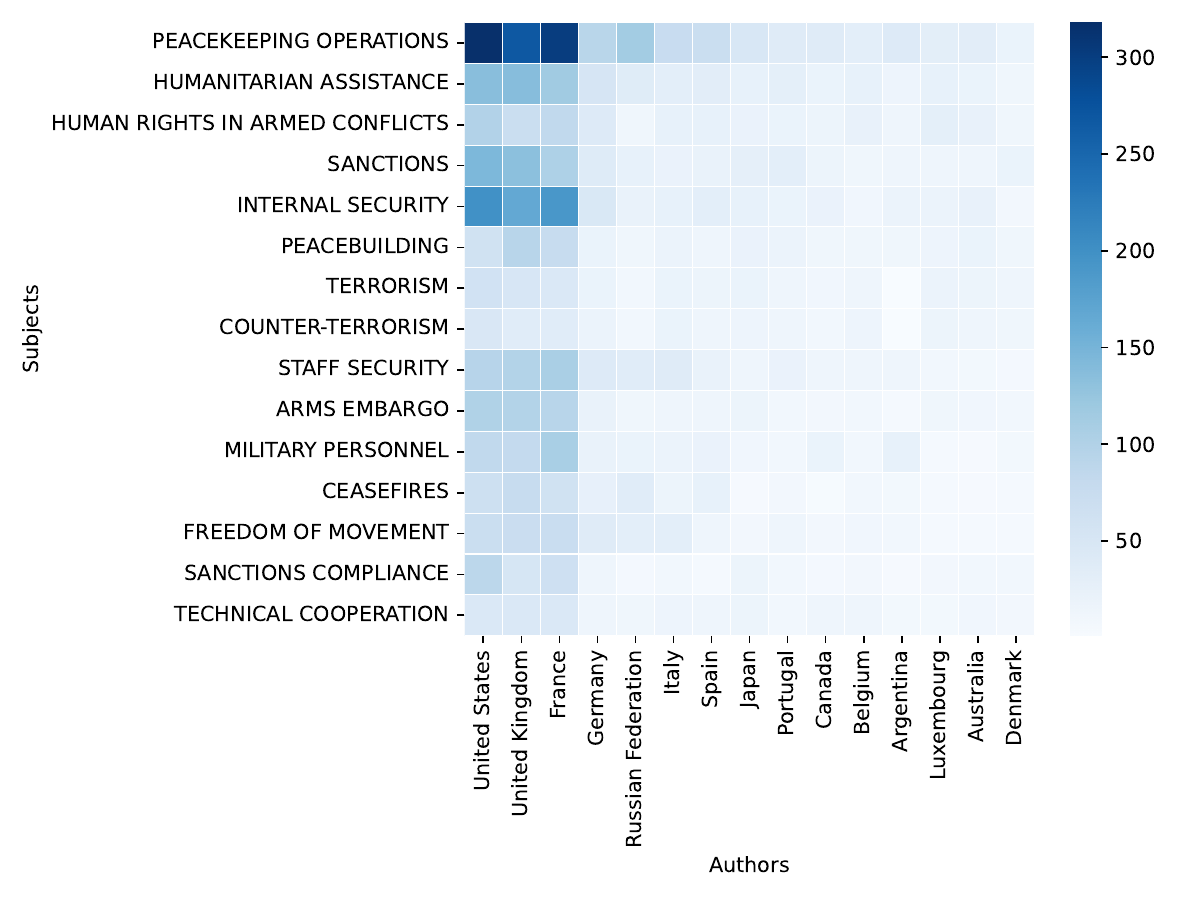}
\caption{\textbf{Author-Subjects Relationships.} This figure shows the co-occurrence matrix of the top 15 authors and subjects. Each cell represents the number of times an author has written about a topic. The darker the cell, the more the author has written about the topic. 
}
\label{fig:author-subject}
\end{figure*}

\subsubsection{Controversial Subjects in UN}

\begin{figure*}[h]
\centering
\includegraphics[width=0.9\linewidth]{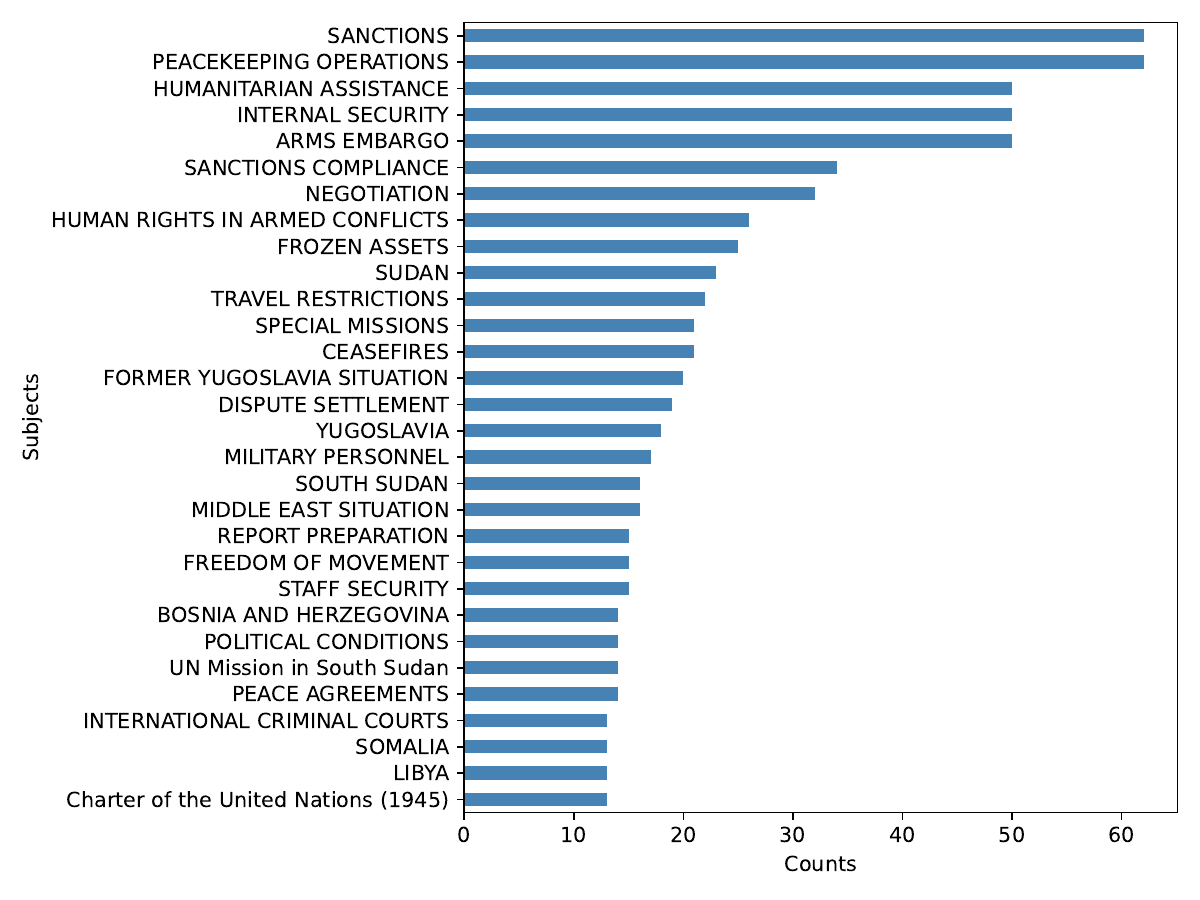}
\caption{This figure shows the top 30 subjects that at least one country did not vote 'Yes' on.}
\label{fig:no_subjects}
\end{figure*}

The author-subjects co-occurrence matrix is shown in Figure~\ref{fig:author-subject}, which reveals distinct patterns in UN Security Council draft resolution authorship. Permanent members of the Security Council—particularly the United States, United Kingdom, and France—demonstrate high engagement across most subject areas, with notably strong involvement in peacekeeping operations and humanitarian assistance. This pattern underscores their important role in global governance while reflecting Western powers' emphasis on human rights and humanitarian interventions. 
From a thematic perspective, peacekeeping operations, humanitarian assistance, and human rights in armed conflicts emerge as the most prominent subjects across member states, forming an interconnected core of Security Council priorities. This pattern suggests a holistic approach to global security challenges, where military peacekeeping efforts are consistently coupled with humanitarian considerations. The strong co-occurrence between sanctions-related topics and peacekeeping operations indicates that the Council frequently employs a dual strategy of enforcement and intervention. Notably, the emergence of terrorism and counter-terrorism as significant themes reflects the evolving nature of global security threats. The data also reveals that technical cooperation and peacebuilding subjects often appear alongside humanitarian assistance, suggesting a long-term approach to crisis resolution that extends beyond immediate security concerns.

The Top-30 subjects that at least one country did not vote 'Yes' is shown in Figure~\ref{fig:no_subjects}.
It reveals that enforcement-related topics—particularly sanctions, peacekeeping operations, and humanitarian assistance—generate the most disagreement in Security Council voting. 
The high frequency of non-affirmative votes on these subjects suggests persistent tensions between international intervention and national sovereignty. Additionally, the presence of international criminal courts and human rights matters among frequently contested subjects highlights the ongoing challenges in balancing international justice with state sovereignty concerns.

\subsubsection{Trend in UN Resolutions}

\begin{figure*}[!hp]
\centering
\includegraphics[width=\linewidth]{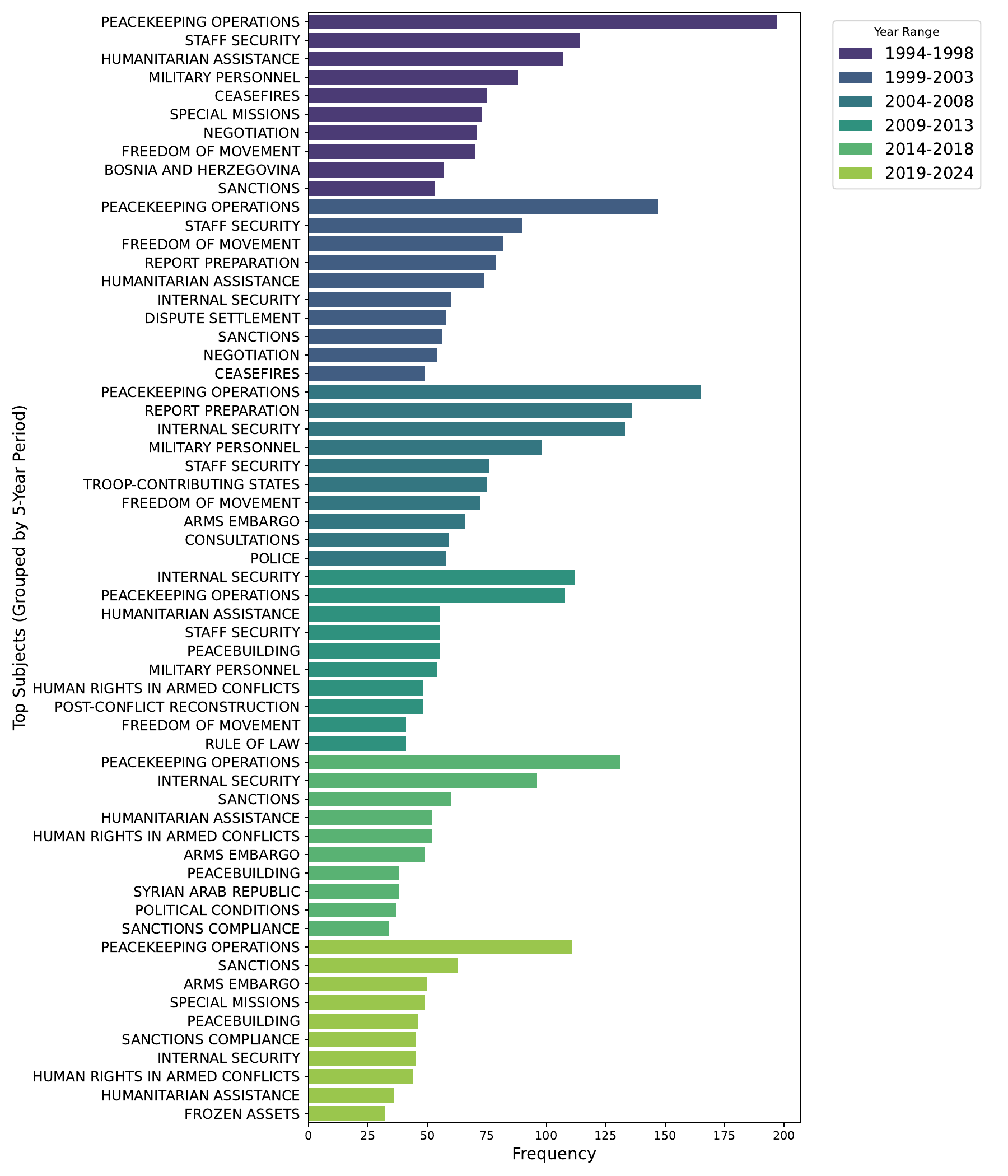}
\caption{This figure shows the top 10 subjects per 5-year period from 1994 to 2024.}
\label{fig:top_10_subjects}
\end{figure*}

\begin{figure}[t]
    \centering
    \includegraphics[width=\linewidth]{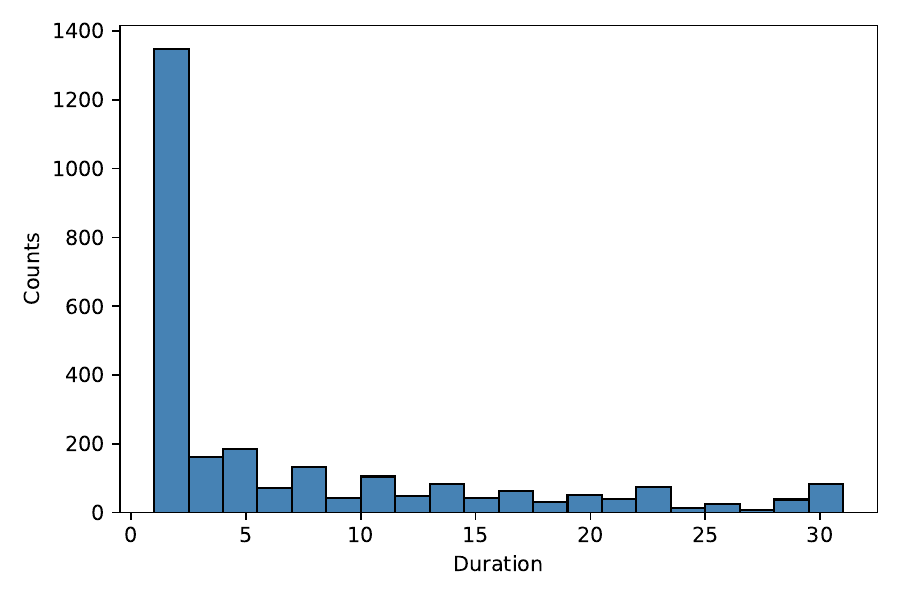}
    \caption{Distribution of the duration of each subject. We can observe that most subjects last for 1 to 5 years, while a few last for more than 30 years.}
    \label{fig:duration}
\end{figure}

The trends in UN resolution topics are also changing over time. The Top-10 subjects per 5-year period from 1994 to 2024 are shown in Figure~\ref{fig:top_10_subjects}. Besides, we also show the distribution of the duration of each subject in Figure~\ref{fig:duration}. Most subjects last for 1 to 5 years, while a few last for more than 30 years.
The two figures reveal two key patterns in the United Nations' focus on global issues. First, certain topics have consistently appeared over the years, indicating the UN’s ongoing attention to these issues. Topics such as international peace and security, human rights, and conflict resolution have remained at the forefront of UN resolutions, reflecting the organization's continuous efforts to address global stability, protect human rights, and resolve conflicts. These persistent topics suggest a sustained, long-term commitment to addressing the most pressing and enduring global challenges. On the other hand, there are topics that have emerged briefly and faded over time, often in response to specific events or crises. For example, resolutions related to regional conflicts or emergency sanctions have been intermittently, typically tied to short-lived geopolitical developments such as military interventions or economic sanctions. These topics highlight the UN's responsive nature, focusing on immediate crises that do not necessarily remain central once the situation is resolved. This pattern underscores the dynamic balance between the UN’s long-standing priorities and its flexibility in addressing emerging global issues as they arise. 
Our benchmark collects all resolutions from this period, making it both challenging and comprehensive, capturing the full scope of the UN's evolving focus on global governance.

\subsection{National-Wised Analysis}

\subsubsection{Voting Frequency}
\begin{figure}[htp]
\centering
\includegraphics[width=\linewidth]{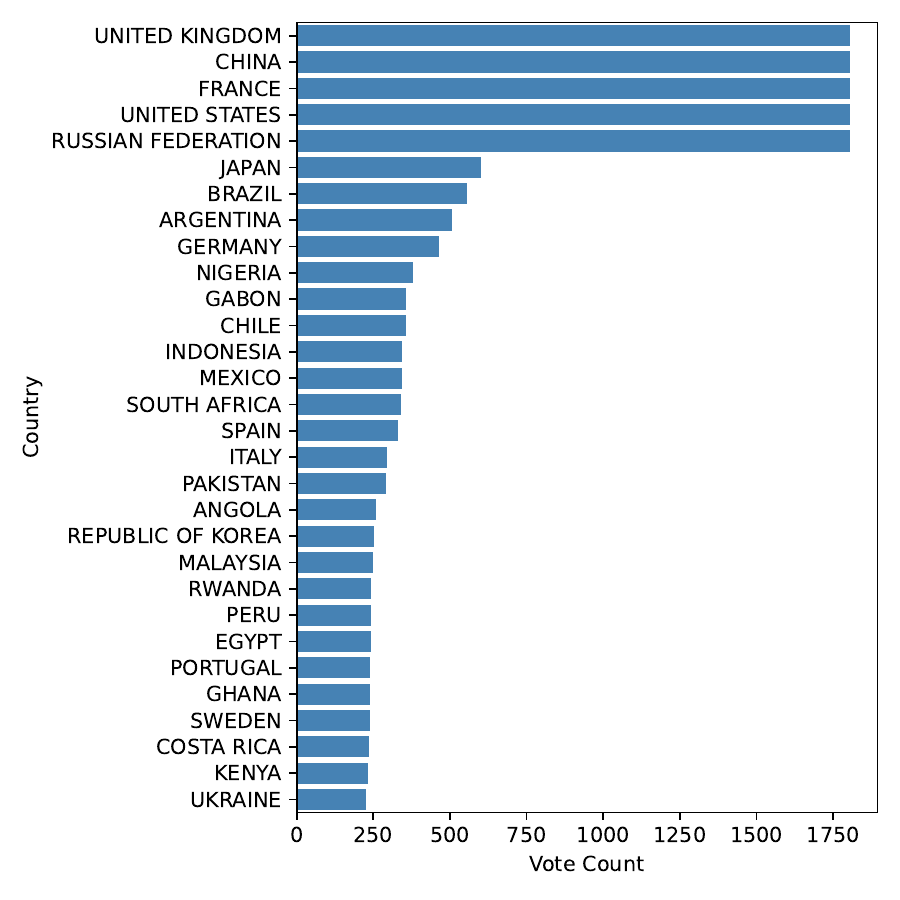}
\caption{The top 30 countries that participated in voting the most.}
\label{fig:vote_frequency}
\end{figure}

Figure~\ref{fig:vote_frequency} shows the vote frequency of Top-30 countries. It reveals a clear dominance by the five permanent members of the UN Security Council (UNSC), namely China, France, Russia, the United Kingdom, and the United States. These countries consistently hold the highest number of votes, reflecting their influential roles in shaping international decisions and maintaining global security. In addition to the permanent members, other non-permanent members such as Japan, Brazil, Argentina, Germany, and Nigeria also appear prominently on the list, highlighting their significant involvement in global governance. Their high voting frequencies may reflect their strategic interests, regional influence, and active participation in international diplomacy. The presence of these countries, along with the permanent members, underscores the UN Security Council's complex decision-making process, where both major powers and key regional players contribute to shaping resolutions and global policies.

\begin{figure}[t]
    \centering
    \includegraphics[width=\linewidth]{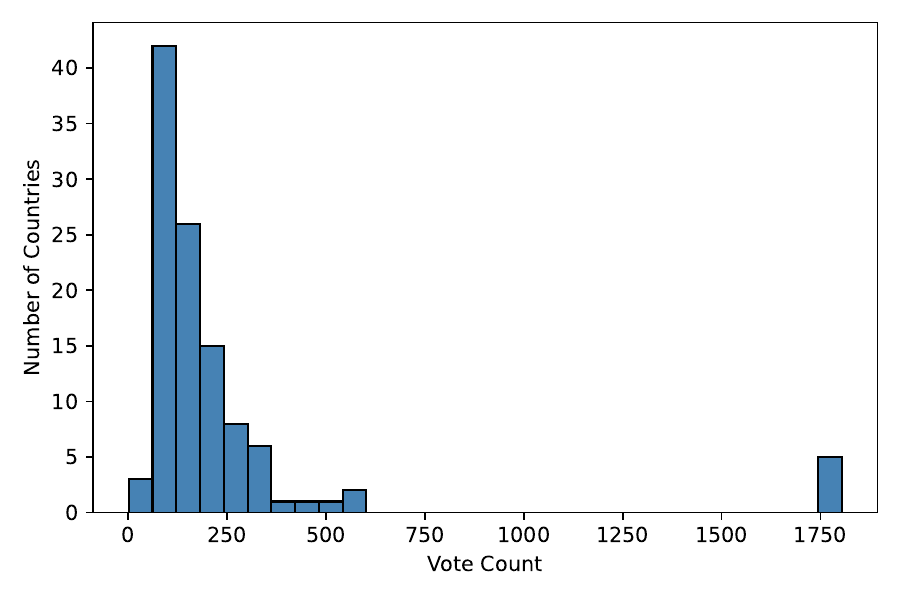}
    \caption{Distribution of the number of votes each country participated}
    \label{fig:resolution frequency}
\end{figure}

The distribution of the number of votes each country participated in, as shown in Figure~\ref{fig:resolution frequency}, reveals that most countries have participated in fewer than 250 votes. This suggests that the majority of countries engage in voting on a limited number of resolutions, likely reflecting their geopolitical priorities and areas of influence within the UN. While some countries may focus on specific regional or issue-based resolutions, others may be more passive in their participation, contributing to fewer votes overall. The relatively small number of countries with more than 250 votes highlights the most active players in UN decision-making, likely including key international powers and nations with significant stakes in global governance. This distribution underscores the varied levels of involvement in the UN’s voting process, with certain countries playing a consistently active role while others engage more selectively.

\begin{table}[h]\centering
\resizebox{0.7\linewidth}{!}{
\begin{tabular}{lrr}\toprule
\textbf{Author} &\textbf{Rejection} &\textbf{Count} \\\midrule
United States &RUSSIAN &68 \\
United States &CHINA &52 \\
United Kingdom &RUSSIAN &47 \\
France &RUSSIAN &45 \\
France &CHINA &39 \\
United Kingdom &CHINA &39 \\
Germany &RUSSIAN &22 \\
Germany &CHINA &21 \\
Japan &CHINA &15 \\
Italy &CHINA &13 \\
\bottomrule
\end{tabular}}
\vspace{5pt}
\caption{This table shows the top 10 pairs of authors and countries that vote not 'Yes' the most. The ``Rejection'' means receiving either a 'No' or 'Abstention' vote. The 'Count' column represents the number of times the author's draft was not voted 'Yes' by the country.}
\label{tab:pair_author_rejection}
\end{table}

\subsubsection{Country Relationships revealed within UN Resolutions}

Table~\ref{tab:pair_author_rejection} presents the top 10 pairs of authors and countries that most frequently voted "No" or abstained on draft resolutions. The data reveals that the United States has the highest number of rejections, with Russia rejecting 68 times and China 52 times. This indicates a consistent divide between the U.S. and these two powers, likely reflecting ongoing geopolitical tensions. Similarly, the United Kingdom and France also show high rejection rates from both Russia and China, suggesting a shared stance among Western powers in opposition to certain resolutions proposed by these countries. On the other hand, countries like Germany, Japan, and Italy appear less frequently in the table, with rejections ranging from 13 to 22 times. These smaller states seem to align more often with the major powers but still demonstrate some differences, particularly with China and Russia. Overall, the table highlights significant diplomatic rifts between the Western powers and Russia/China, with frequent rejections indicating key areas of contention in international relations.

\begin{table}[htp]\centering
\resizebox{0.7\linewidth}{!}{
\begin{tabular}{p{5cm}r}\toprule
\textbf{Country Pair} &\textbf{Count} \\\midrule
(FRANCE, UK) &1,153 \\
(UK, US) &1,147 \\
(FRANCE, US) &1,142 \\
(CHINA, UK) &1,068 \\
(CHINA, FRANCE) &1,064 \\
(CHINA, US) &1,058 \\
(RUSSIAN, UK) &1,035 \\
(FRANCE, RUSSIAN) &1,031 \\
(RUSSIAN, US) &1,024 \\
(CHINA, RUSSIAN) &1,013 \\
\bottomrule
\end{tabular}}
\vspace{5pt}
\caption{This table shows the top 10 pairs of countries that voted 'Yes' together the most. 'US' and 'UK' are the abbreviations for 'UNITED STATES' and 'UNITED KINGDOM'. The 'Count' column represents the number of times the two countries voted 'Yes' together.
}
\label{tab:paired_yes}
\end{table}

Table~\ref{tab:paired_yes} and Table~\ref{tab:paired_no} provide insights into the voting patterns of country pairs within the United Nations, highlighting both strong collaboration and significant divergence in voting behaviors. Table~\ref{tab:paired_yes} shows that certain country pairs, such as France and the United Kingdom (1,153 joint "Yes" votes), United Kingdom and the United States (1,147 joint "Yes" votes), and France and the United States (1,142 joint "Yes" votes), consistently align on many resolutions, reflecting their close diplomatic and strategic cooperation. In contrast, China and the Western powers also exhibit frequent collaboration, with the China-United Kingdom, China-France, and China-United States pairs each voting together over 1,000 times, indicating areas of common interest despite occasional political differences. On the other hand, Table~\ref{tab:paired_no} reveals country pairs that most often did not vote "Yes" together. The China-Russia pair stands out with 69 instances of disagreement, indicating that the two countries share some geopolitical interests. Other pairs, such as Algeria-China (8 times) and Russia-South Africa (6 times), also display divergent voting patterns, reflecting how national and regional interests can influence voting behavior at the UN. These tables underscore the complex dynamics of international diplomacy, where countries may cooperate on certain issues while diverging on others based on their specific interests and priorities.

\begin{table}[htp]\centering
\resizebox{0.7\linewidth}{!}{
\begin{tabular}{p{5cm}r}\toprule
\textbf{Country Pair} &\textbf{Count} \\\midrule
(CHINA, RUSSIAN) &69 \\
(ALGERIA, CHINA) &8 \\
(ALGERIA, RUSSIAN) &7 \\
(RUSSIAN, SOUTH AFRICA) &6 \\
(GABON, RUSSIAN) &6 \\
(CHINA, GABON) &6 \\
(RUSSIAN, VENEZUELA &6 \\
(KENYA, RUSSIAN) &5 \\
(EGYPT, RUSSIAN) &5 \\
(CHINA, INDIA) &5 \\
\bottomrule
\end{tabular}}
\vspace{5pt}
\caption{This table shows the top 10 pairs of countries that did not vote 'Yes' together the most. The 'Count' column represents the number of times the two countries did not vote 'Yes' together.
}
\label{tab:paired_no}
\end{table}

\begin{table}[tp]\centering
\resizebox{\columnwidth}{!}{
\begin{tabular}{lrrrrr}\toprule
&\textbf{2 choices} &\textbf{3 choices} &\textbf{4 choices} &\textbf{5 choices} \\\midrule
Llama-3.2-1B &0.581 &0.394 &0.312 &0.269 \\
Llama-3.2-3B &0.578 &0.393 &0.328 &0.297 \\
Llama-3.1-8B &0.665 &0.507 &0.408 &0.379 \\
GPT-4o &0.726 &0.613 &0.511 &0.464 \\
DeepSeek-V3 &0.695 &0.555 &0.443 &0.422 \\
Mistral-7B &0.563 &0.407 &0.335 &0.281 \\
Qwen2.5-7B &0.642 &0.478 &0.353 &0.293 \\
\bottomrule
\end{tabular}}
\vspace{5pt}
\caption{Comprehensive results for Task 1. }
\label{tab:appendix_task1}
\end{table}

\section{Detailed Results of UNBench}
\label{sec:appendix_results}

In this section, we present the detailed results of the UNBench benchmark, evaluating multiple models across four distinct tasks. The tables provide comprehensive performance metrics for each model on various tasks, including accuracy, precision, recall, AUC, F1 score, and other relevant evaluation metrics.

\begin{table}[tp]\centering
\resizebox{\columnwidth}{!}{
\begin{tabular}{lrrrrr}\toprule
&\textbf{ROUGE} &\textbf{Jaccard} &\textbf{TF-IDF} &\textbf{SentBERT} \\\midrule
Llama-3.2-1B &0.0328 &0.0304 &0.3666 &0.3293 \\
Llama-3.2-3B &0.0407 &0.0341 &0.4287 &0.2902 \\
Llama-3.1-8B &0.0394 &0.0363 &0.4021 &0.3553 \\
GPT-4o &0.1985 &0.1837 &0.7958 &0.6188 \\
DeepSeek-V3 &0.2069 &0.1876 &0.8012 &0.6225 \\
Mistral-7B &0.1935 &0.1688 &0.7522 &0.5750 \\
Qwen2.5-7B &0.2008 &0.1761 &0.7842 &0.6229 \\
\bottomrule
\end{tabular}}
\vspace{5pt}
\caption{Comprehensive results for Task 4. Similarity of IT-IDF and SentBert are calculated by cosine similarity.}
\label{tab:appendix_task4}
\end{table}

Task 1 (Table~\ref{tab:appendix_task1}) evaluates the models' performance across multiple-choice tasks with varying numbers of choices (2 to 5). The results indicate that GPT-4o outperforms the other models across all choice levels, particularly excelling in the 2-choice and 3-choice tasks, where it maintains the highest scores in terms of accuracy. Models like Llama-3.1-8B and DeepSeek-V3 also show competitive results, especially for more complex tasks (4 and 5 choices), though they trail behind GPT-4o.

\begin{table*}[htbp]\centering
\resizebox{\linewidth}{!}{
\begin{tabular}{lcccccccccc}\toprule
&\textbf{Accuracy} &\textbf{AUC} &\textbf{Bal. ACC} &\textbf{Precision} &\textbf{Recall} &\textbf{F1} &\textbf{PR\_AUC} &\textbf{MCC} &\textbf{G-Mean} \\\midrule
Llama-3.2-1B &0.898 &0.497 &0.320 &0.332 &0.320 &0.326 &0.334 &0.006 &0.464 \\
Llama-3.2-3B &0.523 &0.597 &0.597 &0.520 &0.597 &0.402 &0.956 &0.087 &0.597 \\
Llama-3.1-8B &0.917 &0.532 &0.357 &0.360 &0.357 &0.359 &0.338 &0.079 &0.502 \\
GPT-4o &0.922 &0.731 &0.677 &0.400 &0.677 &0.363 &0.343 &0.162 &0.729 \\
DeepSeek-V3 &0.931 &0.720 &0.668 &0.464 &0.668 &0.351 &0.343 &0.151 &0.718 \\
Mistral-7B &0.557 &0.593 &0.426 &0.345 &0.426 &0.268 &0.341 &0.100 &0.569 \\
Qwen2.5-7B &0.935 &0.719 &0.699 &0.373 &0.699 &0.375 &0.344 &0.141 &0.719 \\
\bottomrule
\end{tabular}}
\caption{Comprehensive results for Task 2. }
\label{tab:appendix_task2}
\end{table*}

Task 2 (Table~\ref{tab:appendix_task2}) presents a set of metrics evaluating model performance on binary classification tasks. Here, Qwen2.5-7B leads with the highest accuracy (0.935) and AUC (0.719), indicating its strong ability to differentiate between classes. However, GPT-4o shows superior performance in other metrics such as F1 (0.686) and G-Mean (0.807), making it the most balanced model for this task. DeepSeek-V3 also performs strongly across multiple metrics, especially in precision (0.828), suggesting it excels in tasks where false positives need to be minimized.

\begin{table*}[htp]\centering
\resizebox{\linewidth}{!}{
\begin{tabular}{lcccccccccc}\toprule
&\textbf{Accuracy} &\textbf{AUC} &\textbf{Bal. ACC} &\textbf{Precision} &\textbf{Recall} &\textbf{F1} &\textbf{PR\_AUC} &\textbf{MCC} &\textbf{G-Mean} \\\midrule
Llama-3.2-1B &0.815 &0.546 &0.546 &0.083 &0.245 &0.124 &0.185 &0.057 &0.456 \\
Llama-3.2-3B &0.523 &0.597 &0.597 &0.073 &0.679 &0.132 &0.385 &0.087 &0.591 \\
Llama-3.1-8B &0.935 &0.530 &0.530 &0.211 &0.076 &0.111 &0.168 &0.098 &0.273 \\
GPT-4o &0.968 &0.823 &0.823 &0.714 &0.660 &0.686 &0.696 &0.670 &0.807 \\
DeepSeek-V3 &0.966 &0.724 &0.724 &0.828 &0.453 &0.585 &0.655 &0.597 &0.671 \\
Mistral-7B &0.867 &0.529 &0.529 &0.084 &0.151 &0.108 &0.140 &0.044 &0.370 \\
Qwen2.5-7B &0.926 &0.578 &0.578 &0.250 &0.189 &0.215 &0.241 &0.179 &0.427 \\
\bottomrule
\end{tabular}}
\caption{Comprehensive results for Task 3. }
\label{tab:appendix_task3}
\end{table*}

Task 3 (Table~\ref{tab:appendix_task3}) focuses on multi-class classification tasks, where GPT-4o again stands out with the highest accuracy (0.968) and balanced performance across other metrics like recall, F1, and G-Mean. DeepSeek-V3 shows strong performance in precision (0.828) and recall (0.453), which may indicate a more specialized capability in identifying specific class instances.

Task 4 (Table~\ref{tab:appendix_task4}) assesses the models' ability to generate meaningful representations and comparisons between text using various similarity measures like ROUGE, Jaccard, and cosine similarity. DeepSeek-V3 performs best across multiple metrics, particularly in cosine similarity (0.8012 with TF-IDF and 0.6225 with SentBERT), demonstrating its strength in textual similarity and comparison tasks. GPT-4o also shows strong performance, particularly with cosine similarity (0.7958 with TF-IDF and 0.6188 with SentBERT).

Overall, the results demonstrate the competitive nature of current models, with GPT-4o leading in several tasks due to its balanced performance across various metrics. However, other models like DeepSeek-V3 and Qwen2.5-7B also show strong results in specific areas, such as precision and text similarity. These findings highlight the strengths and limitations of each model, offering valuable insights for selecting the most suitable model for specific tasks within the UNBench framework.

\section{Additional Analyses}
\label{sec:appendix_add}

\subsection{Temporal Robustness Evaluation}
\label{sec:appendix_c1}

To assess the temporal generalization of language models, we divide the test set based on GPT-4o's training cutoff (October 2023) into pre-2023 (seen) and post-2023 (unseen) subsets. This helps us evaluate whether models perform worse when facing emerging, unseen geopolitical developments.

\textbf{Task 2: Voting Simulation.} GPT-4o shows strong performance overall, but drops notably on unseen (post-2023) data. Balanced accuracy falls from 0.675 on seen examples to 0.528 on unseen ones, highlighting the challenge of temporal generalization.

\begin{center}
\begin{tabular}{lcccc}
\toprule
Model & Accuracy & Bal. ACC & AUC & F1 \\
\midrule
GPT-4o         & 0.922 & 0.677 & 0.731 & 0.363 \\
GPT-4o-seen    & 0.924 & 0.675 & 0.727 & 0.360 \\
GPT-4o-unseen  & 0.907 & 0.528 & 0.640 & 0.376 \\
\bottomrule
\end{tabular}
\end{center}

\textbf{Task 3: Draft Adoption Prediction.} In contrast, GPT-4o maintains high accuracy across time splits, suggesting the model’s adoption reasoning is less sensitive to temporal shifts.

\begin{center}
\begin{tabular}{lcccc}
\toprule
Model & Accuracy & Bal. ACC & AUC & F1 \\
\midrule
GPT-4o         & 0.968 & 0.823 & 0.823 & 0.686 \\
GPT-4o-seen    & 0.976 & 0.816 & 0.816 & 0.737 \\
GPT-4o-unseen  & 0.962 & 0.828 & 0.828 & 0.656 \\
\bottomrule
\end{tabular}
\end{center}

\subsection{Geopolitical Group Analysis}
\label{sec:appendix_c2}

We further investigate whether model behavior differs depending on the geopolitical affiliation of draft resolution authors, specifically contrasting Western European and Others Group (WEOG) members with non-WEOG states. This evaluation provides a fairness-oriented lens on LLM behavior.

\textbf{Task 2: Voting Simulation.} GPT-4o performs better on resolutions drafted by non-WEOG countries (balanced accuracy: 0.690) than on those authored by WEOG states (0.537). The discrepancy suggests that the model may struggle to simulate diplomatic reactions to Western-sponsored resolutions.

\begin{center}
\begin{tabular}{lcccc}
\toprule
Group & Bal. ACC & Precision & Recall & PR AUC \\
\midrule
WEOG     & 0.537 & 0.441 & 0.358 & 0.524 \\
Non-WEOG  & 0.690 & 0.384 & 0.690 & 0.346 \\
\bottomrule
\end{tabular}
\end{center}

\textbf{Task 3: Draft Adoption Prediction.} Results are more stable across regions for adoption outcomes, with GPT-4o achieving comparable balanced accuracy on WEOG (0.815) and non-WEOG (0.824) drafts.

\begin{center}
\begin{tabular}{lcccc}
\toprule
Group & Bal. ACC & Precision & Recall & PR AUC \\
\midrule
WEOG     & 0.815 & 0.778 & 0.636 & 0.714 \\
Non-WEOG  & 0.824 & 0.700 & 0.667 & 0.693 \\
\bottomrule
\end{tabular}
\end{center}

\subsection{Contextual Input Variation}
\label{sec:appendix_c3}

We test whether including metadata like dates or authorship improves LLM decision-making. This is especially relevant for interpreting real-world diplomatic context.

\textbf{Task 2: Voting Simulation.} Adding author metadata improves balanced accuracy (from 0.677 to 0.684) and G-Mean (from 0.729 to 0.753), suggesting model predictions benefit from geopolitical signaling embedded in authorship.

\begin{center}
\begin{tabular}{lccc}
\toprule
Setting & Bal. ACC & Recall & G-Mean \\
\midrule
GPT-4o            & 0.677 & 0.677 & 0.729 \\
GPT-4o-dates      & 0.673 & 0.673 & 0.719 \\
GPT-4o-authors    & 0.684 & 0.684 & 0.753 \\
\bottomrule
\end{tabular}
\end{center}

\textbf{Task 3: Draft Adoption Prediction.} Adding date or author metadata boosts both recall and G-Mean, with the strongest performance under the date-enhanced setting.

\begin{center}
\begin{tabular}{lccc}
\toprule
Setting & Bal. ACC & Recall & G-Mean \\
\midrule
GPT-4o            & 0.823 & 0.660 & 0.807 \\
GPT-4o-dates      & 0.888 & 0.811 & 0.885 \\
GPT-4o-authors    & 0.859 & 0.769 & 0.854 \\
\bottomrule
\end{tabular}
\end{center}

\subsection{Model Scaling Analysis}
\label{sec:appendix_c4}

\begin{table*}[!t]
\centering
\begin{tabular}{lcccccccc}
\toprule
&\multicolumn{2}{c}{\textbf{Task 1}} &\multicolumn{2}{c}{\textbf{Task 2}} &\multicolumn{2}{c}{\textbf{Task 3}} &\multicolumn{2}{c}{\textbf{Task 4}} \\
\cmidrule{2-9}
\textbf{Model} & (1/2) & (1/5) &Bal. ACC &PR AUC &Bal. ACC &Mac. F1 &ROUGE &Cosine Sim. \\
\midrule

Llama-3.2-1B &0.581 &0.269 &0.546 &0.185 &0.320 &0.326 &0.033 &0.329 \\

Llama-3.2-3B &0.578 &0.297 &0.597 &0.385 &0.597 &\textbf{0.402} &0.041 &0.290 \\

Qwen2.5-7B & 0.642 & 0.293 & 0.699 & 0.375 & 0.578 & 0.241 & \underline{0.201} & \textbf{0.623} \\

Mistral-7B &0.563 &0.281 &0.426 &0.268 &0.529 &0.140 &0.194 &0.575 \\

Llama-3.1-8B &0.665 &0.379 &0.530 &0.168 &0.357 &0.359 &0.039 &0.355 \\

DeepSeek-V3-70B &\underline{0.695} &\underline{0.422} &\underline{0.724} &\underline{0.655} &\underline{0.668} &0.351 &\textbf{0.207} &\textbf{0.623} \\

GPT-4o (~200B) &\textbf{0.726} &\textbf{0.464} &\textbf{0.823} &\textbf{0.696} &\textbf{0.677} &\underline{0.363} & 0.199 &\underline{0.619} \\

\bottomrule
\end{tabular}
\caption{Model scaling comparison across four tasks. Metrics match those in Table~\ref{tab:exp_major}. Best results are in \textbf{bold}, second-best are \underline{underlined}.}
\label{tab:appendix_model_scaling}
\end{table*}

We compare performance across models of increasing parameter scale to assess how size correlates with capability across tasks. While model size generally aligns with stronger performance, architectural design and training data also have noticeable effects.

Larger models, particularly GPT-4o and DeepSeek-V3, show consistent strength across tasks, with notable gains in political reasoning, coherence, and generation fidelity. However, smaller models like Qwen2.5-7B occasionally match or outperform larger models (e.g., in embedding-based similarity), indicating that scale alone does not guarantee task superiority.

\subsection{Evaluation Strategy Enhancements (Task 4)}
\label{sec:appendix_c5}

To supplement automatic metrics (e.g., ROUGE, cosine similarity), we can incorporate enhanced qualitative assessments like:

\begin{itemize}
    \item \textbf{LLM-as-a-Judge:} GPT-4 evaluates outputs on tone, coherence, justification quality, and alignment with national interests.
    \item \textbf{Expert Review:} Human evaluators from political science backgrounds will provide qualitative ratings and thematic annotations.
\end{itemize}

These strategies offer richer insights beyond n-gram or embedding-based overlap, and will be used to further benchmark models on Task 4 in future iterations of UNBench.

\end{document}